# The Role of CNL and AMR in Scalable Abstractive Summarization for Multilingual Media Monitoring

Normunds Gruzitis and Guntis Barzdins

University of Latvia, IMCS and LETA
normunds.gruzitis@lumii.lv, guntis.barzdins@leta.lv

In the era of Big Data and Deep Learning, a common view is that statistical and machine learning approaches are the only way to cope with the robust and scalable information extraction and summarization. Manning [1] compares Deep Learning with a tsunami at the shores of Computational Linguistics, raising a question if this is the end for the linguistically oriented approaches. Consequently, this question is relevant also to the special interest group on Controlled Natural Language (CNL).

It has been recently proposed that the CNL approach could be scaled up, building on the concept of embedded CNL [2] and, thus, allowing for CNL-based information extraction from e.g. normative or medical texts that are rather controlled by nature but still infringe the boundaries of CNL or the target formalism [3]. It has also been demonstrated that CNL can serve as an efficient and user-friendly interface for Big Data end-point querying [4; 5], or for bootstrapping robust NL interfaces [6]. as well as for tailored multilingual natural language generation from the retrieved data [4].

In this position paper, we focus on the issue of multi-document storyline summarization, and generation of story highlights – a task in the Horizon 2020 Big Data project SUMMA[1] (Scalable Understanding of Multilingual MediA). For this use case, the information extraction process, i.e., the semantic parsing of input texts cannot be approached by CNL: large-scale media monitoring is not limited to a particular domain, and the input sources vary from newswire texts to radio and TV transcripts to user-generated content in social networks. Robust machine learning techniques are necessary instead to map the arbitrary input sentences to their meaning representation in terms of PropBank and FrameNet [7], or the emerging Abstract Meaning Representation, AMR [8], which is based on PropBank with named entity recognition and linking via DBpedia [9]. AMR parsing has reached 67% accuracy (the $F_1$ score) on open-domain texts, which is a level acceptable for automatic summarization [10].

Although it is arguable if CNL can be exploited to approach the robust wide-coverage semantic parsing for use cases like media monitoring, its potential becomes much more obvious in the opposite direction: generation of story highlights from the summarized (pruned) AMR graphs. An example of possible input and expected output is given in Figure 1.

While novel methods for AMR-based abstractive[2] summarization begin to appear [11], full text generation from AMR is still recognized as a future task [11], which is an unexplored niche for the CNL and grammar-based approaches.[3] Here we see, for instance, Grammatical Framework, GF [12], as an excellent opportunity for implementing an AMR to text generator.

---

[1] http://summa-project.eu
[2] Abstractive summarization contrasts extractive summarization which selects representative sentences from the input documents, optionally compressing several sentences into one.

| Article[1] | [..] *An ongoing battle in Aleppo eventually terminated when the rebels took over the city.* [..] *President Assad gave a speech, denouncing the death of soldiers.* [..] |
|---|---|
| Article[2] | [..] *Syrian rebels took control of Aleppo.* [..] |
| Article[3] | [..] *The Syrian opposition forces won the battle over Aleppo city.* [..] *Syrian president announced that such insurgence will not be tolerated.* [..] |

| **Output Summary:** | |
|---|---|
| *Syrian rebels took over Aleppo* <br> Article[1]  Article[2]  Article[3] | *Assad gave a speech about the battle* <br> Article[1]  Article[3] |

**Fig. 1.** Abstractive summarization. An example from the SUMMA proposal

The summarized AMR graphs would have to be mapped to the abstract syntax trees (AST) in GF (see an example in Figure 2). As GF abstract syntax can be equipped with multiple concrete syntaxes, reusing the readily available GF resource grammar library, this would allow for multilingual summary generation, even extending the SUMMA proposal.

| **Abstract Meaning Representation (AMR)** | **GF Abstract Syntax Tree (AST)** |
|---|---|
| ```
(x3 / take-01
 :ARG0 (x2 / organization
  :wiki "Syrian_opposition"
  :name (n2 / name
   :op1 "Syrian" :op2 "rebels"))
 :ARG1 (x5 / city :wiki "Aleppo"
  :name (n / name :op1 "Aleppo")))
``` | ```
PredVP
  organization_Syrian_opposition_NP
  (ComplV2
    take_over_V2
    city_Aleppo_NP)
``` |

**Fig. 2.** AMR and AST representations of "*Syrian rebels took over Aleppo*"

We assume that the generation of story highlights in the open newswire domain is based on a relatively limited set of possible situations (semantic frames) and a relatively limited set of syntactic constructions (a restricted style of writing) similar to the Multilingual Headlines Generator demo[4] illustrated in Figure 3.

Although we have not yet implemented a method for the automatic mapping of AMR graphs to AST trees, there is a clear relation between the two representations. From the CNL / GF perspective, the main issue is the open lexicon (named entities and their translation equivalents), however, the AMR *wiki:* links to DBpedia would enable the acquisition of a large-scale multilingual GF lexicon of named entities (as implicitly illustrated in Figure 2).

---

[3] SemEval-2017 is expected to host a competition on AMR to text generation.
[4] http://www.grammaticalframework.org/demos/multilingual_headlines.html

| Basque | Portuguese | French |
|---|---|---|
| Atzerri ministroak akordioa sinatuko du | O ministro dos negócios estrangeiros assinará o acordo | Le ministre des affaires étrangères signera l'accord |
| **Italian** | **Romanian** | **English** |
| Il ministro degli affari esteri firmerà l'accordo | Ministrul afacerilor externe va semna acordul | The minister of foreign affairs will sign the agreement |
| **German** | **Swedish** | **Latvian** |
| Der Außenminister wird das Abkommen unterzeichnen | Utrikesministern ska skriva på överenskommelsen | Ārlietu ministrs parakstīs vienošanos |

**Fig. 3.** Multilingual Headlines Generator implemented in GF by José P. Moreno

With the technique outlined in this paper, the simplified Multilingual Headlines Generator would effectively become the Multilingual Headlines Summarizer with wide applicability in the SUMMA project and beyond.

**Acknowledgements**

This work is supported in part by the H2020 project SUMMA (under grant agreement No. 688139), and by the Latvian state research programmes NexIT and SOPHIS.